\documentclass[sigconf]{acmart}

\usepackage{booktabs}
\usepackage{dblfloatfix}
\usepackage{lmodern}
\usepackage{balance}

\setcopyright{none}

\begin{document}
\settopmatter{printacmref=false}
{\fontfamily{phv}\selectfont
\title{Feature selection of neural networks is skewed\\ towards the less abstract cue}
\settopmatter{printacmref=false}

\author{Marcell Wolnitza}
\authornote{Corresponding author.~Marcell Wolnitza is also with Georg-August-University G\"ottingen,  Friedrich-Hund-Platz 1, 37077 G\"ottingen, Germany.}
\affiliation{%
  \centering
  \institution{University~of~Applied~Sciences~Koblenz}
  \institution{Joseph-Rovan-Allee 2}
  \institution{53424 Remagen, Germany}
  \Large
  }
\email{wolnitza@hs-koblenz.de}

\author{Babette Dellen}
\affiliation{%
  \centering
  \institution{University~of~Applied~Sciences~Koblenz}
  \institution{Joseph-Rovan-Allee 2}
  \institution{53424 Remagen, Germany}
  \Large
  }
\email{dellen@hs-koblenz.de}
\email{}\email{}\email{}
}

\begin{abstract}
Artificial neural networks (ANNs) have become an important tool for image classification with many applications in research and industry. However, it remains largely unknown how relevant image features are selected  and how  data properties affect this process. In particular, we are interested whether the abstraction level of image cues correlating with class membership influences feature selection. We perform experiments with binary images that contain a combination of cues, representing two different levels of abstractions: one is a pattern drawn from a random distribution where class membership correlates with the statistics of the pattern, the other a combination of  symbol-like entities, where the symbolic code correlates with class membership. When the network is trained with data in which both cues are equally significant, we observe that the cues at the lower abstraction level, i.e., the pattern, are learned, while the symbolic information is largely ignored, even in networks with many layers. Symbol-like entities are only learned if the importance of low-level cues is reduced compared to the high-level ones. These findings raise important questions about the relevance of features that are learned by deep ANNs and how learning could be shifted towards symbolic features.
\end{abstract}

\keywords{Neural Networks, Pattern Recognition, Feature Selection}
\maketitle

\section{Introduction}
\label{sec:introduction}

\begin{figure}[h]
\centering
\includegraphics[width=\columnwidth]{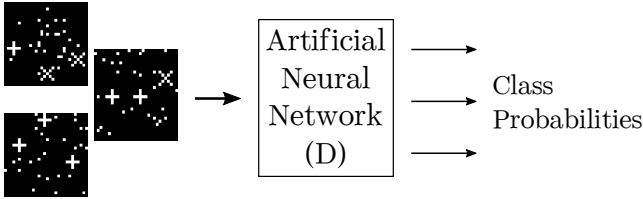}
\caption{Depiction of the training scenarios. We train a neural network for each dataset: {\small$\mathbf{Both~Cues^{train}~(A)}$}, {\small$\mathbf{Symbol^{train}~(B)}$}, {\small$\mathbf{Pattern^{train}~(C)}$} and {\small$\mathbf{Dist.~Both~Cues^{train}~(D)}$}.}
\label{fig:idea}
\end{figure}

\begin{figure}[h]
\centering
\includegraphics[width=\columnwidth]{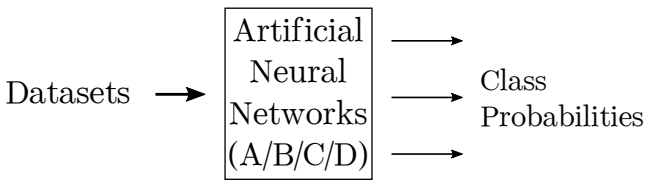}
\caption{Depiction of the test procedures. We evaluate the prediction capability  of the class probabilities of every previously trained neural network on all test-data subsets: {\small$\mathbf{Both~Cues^{test}}$}, {\small$\mathbf{Symbol^{test}}$}, {\small$\mathbf{Pattern^{test}}$} and {\small$\mathbf{Dist.~Both~Cues^{test}}$}.}
\label{fig:idea_test}
\end{figure}

Advances in the manufacturing of fast graphic processing units (GPU) and the availability of large datasets have permitted the successful use of neural networks in many diverse fields, such as natural-language processing \citep{dlspeech} or  computer vision \citep{cv}. State-of-the-art results are obtained by deep learning \citep{dlnature}, which utilizes the large capacity of many layers for representing non-linear functions that map the input signal to an output. However, there is little profound theoretical insight   how the learning process is affected by the data, and which input features are actually encoded by the network and where \citep{blackbox}. \\
A common explanation of data processing in deep neural networks focuses on an interpretation of the hidden layers \citep{dlnature, layerunderstanding}. Early layers extract low-level features from the raw input data, which are further combined in the middle and last layers to obtain a high-level representation. For example, in object classification the first layers in the network extract edge features. These are combined into general object parts and finally assembled to ``archetype objects'' \citep{dlnature, visualizefeatures}. This processing pipeline exhibits similarities with presumed biological neural processing strategies \citep{neuroscience}. Complementary, in learning theory \citep[e.g.][]{vapnik} one tries to formulate general mathematical principles of the system, from which important properties, such as learning boundaries, consistency of the learning process, generalization capabilities, are derived, including the representational power of deep neural networks \citep{reppower1, reppower2}. A theoretical framework to examine (deep) neural networks is the so called information bottleneck \citep{infobottleneck}, which makes general claims about the relevant information contained in random input variables with respect to random output variables and the network's optimization via stochastic gradient descent. However, there is ongoing research and discussion about the applicability of this theory \citep{ibbad, ibdiscussion}.\\
In this paper, we present an experimental setup how the learning process in neural networks, i.e., feature selection, is affected by the input data. We create synthetic, binary images containing two cues, which differ in their level of abstraction and define three distinct classes. The low-level cue is a pattern drawn from a random distribution, being different for every class. The high-level cue is a combination of three symbols occurring in the image according to a class-specific code. We use datasets that are made of these images and the respective class labels to train an ANN to perform the classification task (see Figure~\ref{fig:idea}). We then evaluate the classification performance of the trained networks for every test set (see Figure~\ref{fig:idea_test}). Every cue by itself suffices to correctly classify an image.\\ 
While most approaches concentrate on exploring the network explicitly in terms of interpretation of layers or its capability to represent highly non-linear functions, our study focuses on the data: If two kinds of cues to classify the data are presented to the network, will a combination of them be used or only a single one? And, if the latter holds true, which type of cue is being favored, and are we able to influence the decision made by the network?

\newpage
\section{Material and methods} 
\label{sec:material_methods}

In this work we investigate which kind of cues are used by simple feed-forward neural networks to learn from images. For this purpose we choose a standard neural network architecture (see Section~\ref{subsubsec:architecture}) and create sets of synthetic images containing different kind of cues, each of them correlating with class membership (see Section~\ref{subsubsec:images}).

\subsection{Synthetic Datasets}
\label{subsec:synth_datasets}

For our experiments we use four datasets, each of them including $30,000$ images. Every image contains a combination of high- and low-level cue or only a single cue. Every cue is class-specific. The datasets differ in the type of cue used in the images. In Section~\ref{subsubsec:images} we describe general properties of the images and explain the composition of the cues and classes in more detail. In Section~\ref{subsubsec:datasets} we describe the datasets.

\subsubsection{Images}
\label{subsubsec:images}

\begin{description}
	\item[General properties] The images for the experiments have a size of $30$~x~$30$ pixels and are binary. We create them synthetically with the Software MATLAB. The amount of pixels used to create a single cue is always $27$ and independent of the type of cue. See Figure~\ref{fig:idea} on the left for examples.
	\item[Cues] The cues we use to create the images correlate with class membership. We differentiate between high- and low-level cues that differ in their complexity. We define complexity as the number of iterations needed for the neural network to classify the feature correctly. In this case higher level means that more iterations are needed (see Section~\ref{sec:conclusion}). For the high-level cue we use a combination of three symbols (``+'' and ``x''). Every single symbol is made of nine pixels to ensure scale invariance and placed uniform randomly in the image.\\For the low-level cue we use a pattern of pixels, drawn from a random distribution. In Figure~\ref{fig:idea} samples from the different datasets are shown.
	\item[Classes] In this experiment we construct three different classes: The first class correlates with a  uniformly-\linebreak distributed pattern as the low-level cue. It also correlates with a combination of two symbols ``x'' and one ``+'' as the high-level cue or a combination of them (depending on the dataset used). The second class correlates with a pattern that is drawn from a distribution accumulating the pixels in the center of the image. It also correlates with a combination of two symbols ``+'' and one ``x'' or their combination. The third class correlates with a pattern that is drawn from a distribution accumulating the pixels in the corner of the image and also with three symbols ``+'' or the combination of both cues. See Table~\ref{tab:classes} for a comprehensive overview of the definitions of the classes.
\end{description}

\begin{table}[!h]
\centering
\caption{Composition of the classes in datasets {\small$\mathbf{Both~Cues}$}, {\small$\mathbf{Symbol}$} and {\small$\mathbf{Pattern}$}.}
\label{tab:classes}
\begin{tabular}{ccc}
\textbf{Class} & \textbf{pattern distribution} & \textbf{symbols}\\
\hline \hline
I & uniform & +xx\\
\hline
II & centered & ++x\\
\hline
III & cornered & +++\\
\hline
\end{tabular}
\end{table}

\newpage
\subsubsection{Datasets}
\label{subsubsec:datasets}

\begin{description}
	\item[Both Cues] This dataset contains $30,000$ images with both, high- and low-level cues together and their corresponding class labels. For every class we create $10,000$ images. We split the whole dataset into $22,500$ images ($75~\%$) for the train set and $7,500$ images ($25~\%$) for the test set. We name the train dataset $Both~Cues^{train}$ and the test dataset $Both~Cues^{test}$. In Figure~\ref{fig:idea} examples for dataset $Both~Cues$ are shown.
	\item[Symbol] The second dataset contains $30,000$ images with only the high-level cue present and corresponding labels. For every class we create $10,000$ images, as well. We split the whole dataset into $22,500$ images ($75~\%$) for the train set and $7,500$ images ($25~\%$) for the test set. We name the train dataset $Symbol^{train}$ and the test dataset $Symbol^{test}$. In Figure~\ref{fig:idea} examples for dataset $Symbol$ are shown.
	\item[Pattern] This dataset contains $30,000$ images with only the low-level cue and their corresponding class labels. For every class we create $10,000$ images. We split the whole dataset into $22,500$ images ($75~\%$) for the train set and $7,500$ images ($25~\%$) for the test set. We name the train dataset $Pattern^{train}$ and the test dataset $Pattern^{test}$. In Figure~\ref{fig:idea} examples for dataset $Pattern$ are shown.
	\item[Dist. Both Cues] This dataset contains $30,000$ images with a combination of both, high- and low-level cues and labels as in dataset $Both~Cues$. However, we dilute the dataset by intentionally providing false labels to $23~\%$ of the samples for the pattern. This leads to smaller correlations of the pattern with class membership compared to the correlation of the symbols. Doing so, we want to trigger a different learning behavior of the network than using dataset $Both~Cues$. Apart from that, we proceed like with the other datasets and split the set into $22,500$ images ($75~\%$) for training and $7,500$ images ($25~\%$) for testing, obtaining the datasets $Dist.~Both~Cues^{train}$ and $Dist.~Both~Cues^{test}$. In Figure~\ref{fig:idea} examples for dataset $Dist.~Both~Cues$ are shown and Table~\ref{tab:wrong_classes} provides an overview of the classes we assigned to this manipulated dataset.
\end{description}

\begin{table}[!t]
\centering
\caption{Composition of the distorted classes we are using in {\small$\mathbf{23~\%}$} of the dataset {\small$\mathbf{Dist.~Both~Cues}$}. The rest of the dataset is created as described in Table \ref{tab:classes}.}
\label{tab:wrong_classes}
\resizebox{\columnwidth}{!}{
\begin{tabular}{ccc}
distorted class & pattern distribution & symbol combination\\
\hline \hline
I & centered or cornered & +xx\\
\hline
II & cornered or uniform & ++x\\
\hline
III & uniform or centered & +++\\
\hline
\end{tabular}}
\end{table}
 
\subsection{Neural Network}
\label{subsec:NN}
\subsubsection{Architecture}~
\label{subsubsec:architecture}
\begin{figure}[!h]
\centering
\includegraphics[width=.75\columnwidth]{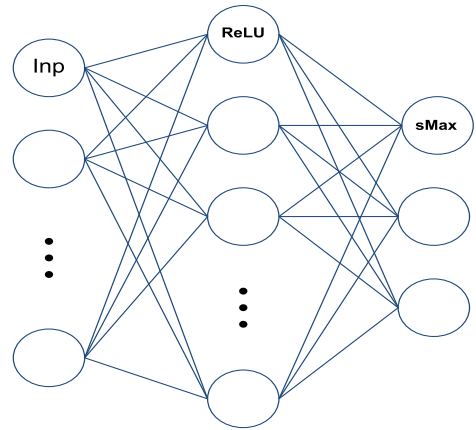}
\caption{Neural network architecture with one hidden layer. The hidden neurons use the ReLU activation function and the output neurons use the Softmax (sMax) activation function for classification. We use setups with two, three and {\small$\mathbf{10}$} hidden layers in this work, too.}
\label{fig:architecture}
\end{figure}

\linebreak
For the experiments we use simple, feed-forward, fully-\linebreak connected neural networks. The architecture of a neural network with one hidden layer is shown in Figure~\ref{fig:architecture}.\\
The pictures used for our investigations have a size of $30$~x~$30$ pixels and are flattened for the input layer. We use one, two, three or $10$ hidden layers, with either $10$, $100$ or $500$ neurons to check the influence of the number of hidden layers and neurons on the resulting test accuracies. All hidden neurons use the ReLU-activation function. Because we are investigating a classification problem, we have three output neurons (for the three different classes). We use a Softmax activation function to output the corresponding probabilities. The results for every setup of the networks are reported in Appendix A.

~
\subsubsection{Training}~
\label{subsubsec:training}
\linebreak
For every training scenario we use $22,500$ images per dataset. We use the batch gradient descent method with $32$ randomly chosen images per iteration and a learning rate of $1~\cdot~10^{-3}$. The error is calculated as the cross entropy of the network output and the provided labels. To avoid over-fitting on the train set, we limit the maximum number of epochs to $1000$ and implement early stopping. One epoch is a complete run over the whole dataset. To test the influence of the number of hidden neurons and layers we execute runs with $10$, $100$ or $500$ hidden neurons and one, two, three or $10$ hidden layers, respectively.\\
We implement and run our experiments with the Keras API of the GPU-accelerated version $1.10$ of Tensorflow \citep{tensorflow}.

\section{Results and Discussion}
\label{sec:results}

We investigate many different training and testing scenarios in this work (see Figures~\ref{fig:idea} and \ref{fig:idea_test}). To keep this section clear we will therefore organize the results in two parts, namely experiment A and experiment B, and go through them step by step.\\In experiment A we consider only datasets $Both~Cues$,\linebreak$Symbol$ and $Pattern$ to investigate the neural networks behavior. Both cues are equally present. By doing so, we measure the performance of the network and identify the cue that was used by the network to predict the classes.\\In the second experiment B, we use dataset $Dist.~Both~Cues$ that is constructed to contain false low-level cues for $23~\%$ of the samples (see Section~\ref{subsubsec:datasets}). We train the neural network on the subset $Dist.~Both~Cues^{train}$ and check the test accuracy on every dataset $Both~Cues^{test}$, $Symbol^{test}$, $Pattern^{test}$ and $Dist.~Both~Cues^{test}$.\\In Tables~\ref{tab:all_results1} and \ref{tab:all_results2} of the Appendix A a complete overview of the mean test accuracies for all training and test scenarios in every setup of the neural networks is shown. In Table~\ref{tab:sym_overview} a comprehensive overview over the classification capability is presented. All reported results are mean values and mean errors of the mean over five runs.

\begin{table}[h!]
\caption{Symbolic overview for the overall classification performance of the trained networks A, B, C and D on all test subsets. The symbols $\surd$ indicate a mean test accuracy of~{\small$\mathbf{~>~90~\%}$}, {\large $\times$} {\small$\mathbf{<~40~\%}$} and $\bigcirc$ between {\small$\mathbf{40~\%}$} and {\small$\mathbf{90~\%}$}.}
\label{tab:sym_overview}
\resizebox{\columnwidth}{.155\textheight}{
\begin{tabular}{ccc}
training subset & testing subset & classification performance\\
\hline \hline
& $Both~Cues^{test}$ & $\surd$\\
$Both~Cues^{train}$ & $Symbol^{test}$ & {\LARGE $\times$}\\
(Network A) & $Pattern^{test}$ & $\surd$\\
& $Dist.~Both~Cues^{test}$ & $\bigcirc$\\ \hline
& $Both~Cues^{test}$ & $\bigcirc$\\
$Symbol^{train}$ & $Symbol^{test}$ & $\surd$\\
(Network B) & $Pattern^{test}$ & {\LARGE $\times$}\\
& $Dist.~Both~Cues^{test}$ & $\bigcirc$\\ \hline
& $Both~Cues^{test}$ & $\bigcirc$\\
$Pattern^{train}$ & $Symbol^{test}$ & {\LARGE $\times$}\\
(Network C) & $Pattern^{test}$ & $\surd$\\
& $Dist.~Both~Cues^{test}$ & $\bigcirc$\\ \hline
& $Both~Cues^{test}$ & $\surd$\\
$Dist.~Both~Cues^{train}$ & $Symbol^{test}$ & $\bigcirc$\\
(Network D) & $Pattern^{test}$ & $\bigcirc$\\
& $Dist.~Both~Cues^{test}$ & $\bigcirc$\\
\end{tabular}}
\end{table}

\subsection{Experiment A}
\label{subsec:results_A}

\begin{table}[!h]
\centering
\caption{Representative mean test accuracies for one hidden layer with {\small$\mathbf{100}$} hidden neurons of the trained networks.}
\label{tab:first_results}
\resizebox{\columnwidth}{!}{
\begin{tabular}{cccc}
data subset & $Both~Cues^{test}$ & $Symbol^{test}$ & $Pattern^{test}$\\
\hline \hline
$Both~Cues^{train}$ & \textbf{97.46 $\pm$ 0.04} & \textit{33.26 $\pm$ 0.17} & \textbf{96.17 $\pm$ 0.08}\\
$Symbol^{train}$ & 67.93 $\pm$ 0.71 & \textbf{100 $\pm$ 0.00} & \textit{38.69 $\pm$ 0.30}\\
$Pattern^{train}$ & 79.18 $\pm$ 0.25 & \textit{33.48 $\pm$ 0.25} & \textbf{99.25 $\pm$ 0.02}\\
\end{tabular}}
\end{table}

Table~\ref{tab:first_results} shows representatively the accuracies on the test sets after training on the corresponding training sets for neural networks with one hidden layer and $100$ hidden neurons. Using other setups with a different number of hidden layers or hidden neurons did not impact the key findings significantly. We will discuss the effect on the absolute accuracies in Section~\ref{subsec:influence}.\\The best accuracies are obtained when the training and test dataset contain the same kind of cue (main diagonal). This is an expected result, because the training and test subsets are from the same datasets. The worst accuracies are obtained for the cases where we train on the subset $Symbol^{train}$ (or $Pattern^{train}$) and test on $Pattern^{test}$ (or $Symbol^{test}$). With three possible classes an accuracy of around $33~\%$ corresponds to guessing. This means that the network is not able to infer the classes correctly.\\When training the neural network on only $Symbol^{train}$ or $Pattern^{train}$ and evaluating on $Both~Cues^{test}$, we obtain $67.93~\%$ (or $79.18~\%$) test accuracy. This effect is bigger for training on the symbols than for training on the pattern.\\Next, the network is trained for the case of learning with $Both~Cues^{train}$ and tested on $Symbol^{train}$ and $Pattern^{train}$, respectively. While the ANN is able to classify the statistical pattern with $96.17~\%$ test accuracy, only $33.26~\%$ test accuracy is obtained for the symbols.\\The results suggests that the ANN learns only the low-level cue when both cues are being provided together.

\subsection{Experiment B}
\label{subsec:result_B}

Now we evaluate the performance for the additional dataset $Dist.~Both~Cues$, which contains wrong patterns for $23~\%$ of the samples (see Section~\ref{subsubsec:datasets}). Our intention is to trigger a different learning behavior, because the network should not be able to classify the data completely by using only the low-level cue. In Table~\ref{tab:dist_results}, the results are presented.\\The test accuracies for training on the $Both~Cues^{train}$ subset and testing on the $Dist.~Both~Cues^{test}$ confirm the results of the first experiment A. They are in a range of around $76~\%$, which agrees with the percentage of correct patterns. We conclude that the network uses only the low-level cue in this case.\\The second observation is that using the $Dist.~Both~Cues^{train}$ subset for training influenced the learning behavior of the neural network compared to the previous experiment. The test accuracies on the $Both~Cues^{test}$ subset decreased compared to training with the $Both~Cues^{train}$ subset. More importantly, the test accuracies for $Symbol^{test}$ and $Pattern^{test}$ also changed. This indicates that the network now learns both cues. Another indication are the test accuracies for $Dist.~Both~Cues^{test}$ ($80.41~\%/87.09~\%/86.53~\%$) after training on $Dist.~Both~Cues^{train}$. They are above the $23~\%$ error rate of the pattern, indicating that the network does not rely solely on them.\\

\begin{table*}[!t]
\caption{Mean test accuracies for different training and testing scenarios of neural networks with one hidden layer and {\small$\mathbf{10}$}, {\small$\mathbf{100}$} and {\small$\mathbf{500}$} hidden neurons.}
\label{tab:dist_results}
\begin{tabular}{cccccc}
\# hidden neurons & data subset & $Both~Cues^{test}$ & $Symbol^{test}$ & $Pattern^{test}$ & $Dist.~both~cues^{test}$\\
\hline \hline \\
10 & $Both~Cues^{train}$ & \textbf{97.42 $\pm$ 0.03} & \textit{33.01 $\pm$ 0.15} & \textbf{95.81 $\pm$ 0.14} & 76.06 $\pm$ 0.31 \\
& $Symbol^{train}$ & 61.17 $\pm$ 1.50 & \textbf{99.75 $\pm$ 0.09} & \textit{37.86 $\pm$ 0.63} & 60.90 $\pm$ 1.61 \\
& $Pattern^{train}$ & 78.73 $\pm$ 0.19 & \textit{33.54 $\pm$ 0.26} & \textbf{99.25 $\pm$ 0.02} & 63.93 $\pm$ 0.24 \\
& $Dist.~Both~Cues^{train}$ & \textbf{92.63 $\pm$ 0.40} & 73.31 $\pm$ 1.76 & 81.22 $\pm$ 0.92 & 80.41 $\pm$ 0.99 \\
\\
100 & $Both~Cues^{train}$ & \textbf{97.46 $\pm$ 0.04} & \textit{33.26 $\pm$ 0.17} & \textbf{96.17 $\pm$ 0.08} & 75.61 $\pm$ 0.12 \\
& $Symbol^{train}$ & 67.93 $\pm$ 0.71 & \textbf{100 $\pm$ 0.00} & \textit{38.69 $\pm$ 0.30} & 67.95 $\pm$ 0.45 \\
& $Pattern^{train}$ & 79.18 $\pm$ 0.25 & \textit{33.48 $\pm$ 0.25} & \textbf{99.25 $\pm$ 0.02} & 64.33 $\pm$ 0.12 \\
& $Dist.~Both~Cues^{train}$ & \textbf{96.31 $\pm$ 0.16} & 85.01 $\pm$ 0.53 & 77.38 $\pm$ 0.14 & 87.09 $\pm$ 0.12 \\
\\
500 & $Both~Cues^{train}$ & \textbf{97.59 $\pm$ 0.03} & \textit{33.28 $\pm$ 0.18} & \textbf{96.54 $\pm$ 0.13} & 75.92 $\pm$ 0.29\\
& $Symbol^{train}$ & 66.33 $\pm$ 0.35 & \textbf{100 $\pm$ 0.00} & \textit{39.10 $\pm$ 0.30} & 65.91 $\pm$ 0.16 \\
& $Pattern^{train}$ & 79.10 $\pm$ 0.23 & \textit{33.35 $\pm$ 0.25} & \textbf{99.19 $\pm$ 0.02} & 63.89 $\pm$ 0.13 \\
& $Dist.~Both~Cues^{train}$ & \textbf{97.31 $\pm$ 0.10} & 86.45 $\pm$ 0.58 & 78.53 $\pm$ 0.38 & 86.53 $\pm$ 0.19 \\
\end{tabular}
\vspace*{.33cm}
\end{table*}

\subsection{Influence of the Number of Hidden Neurons and Layers on the Performance}
\label{subsec:influence}

\begin{figure}[!h]
\centering
\includegraphics[width=\columnwidth]{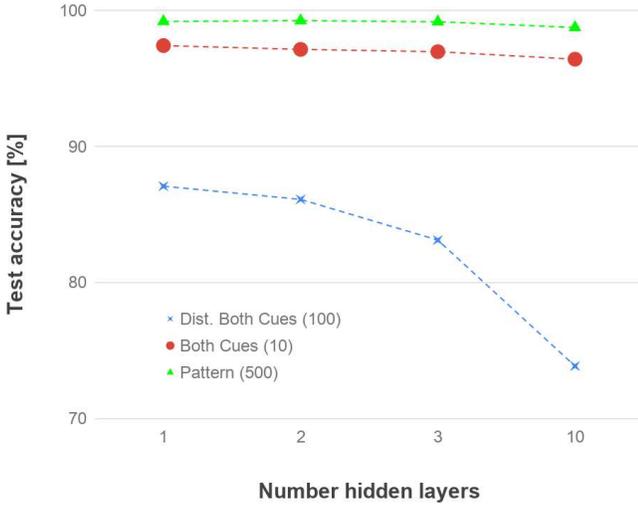}
\caption{Examples of decaying test accuracies by adding more hidden layers. The number of hidden neurons per layer is indicated in brackets.}
\label{fig:decay}
\end{figure}

\begin{figure}[!h]
\centering
\includegraphics[width=\columnwidth]{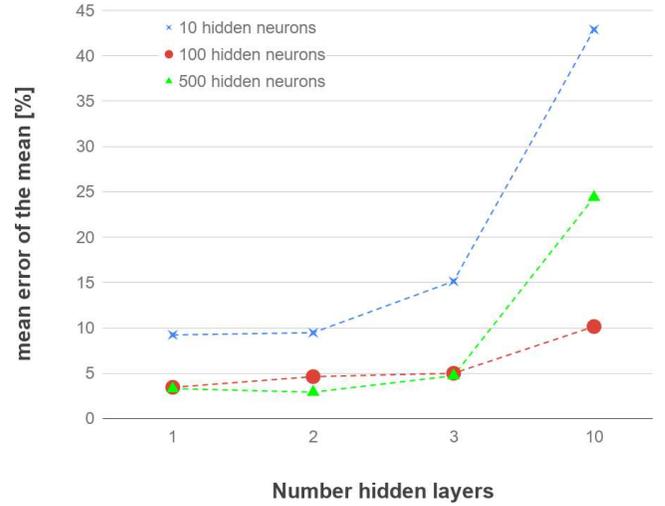}
\caption{Examples of the increase of the mean error of the mean by adding more hidden layers for {\small$\mathbf{10}$}, {\small$\mathbf{100}$} and {\small$\mathbf{500}$} hidden neurons per layer.}
\label{fig:meom}
\end{figure}

We repeated the experiments for ANN with different numbers of hidden neurons and layers. The results are presented in Appendix A. In general, similar behavior is observed, though some differences can be found. Using batch normalization did not have a significant impact on the general trend. Figures~\ref{fig:decay} and \ref{fig:meom} show the slight decrease of the test accuracies and increase of the mean errors of the mean by using more hidden layers.

\begin{table*}[b]
\centering
\caption{Mean number of epochs over five runs needed for convergence during training.}
\label{tab:iterations}
\begin{tabular}{cc}
training dataset & \# epochs\\
\hline \hline
$Both~Cues$ & 45.8 $\pm$ 0.8\\
\hline
$Symbol$ & 849.0 $\pm$ 115.45\\
\hline
$Pattern$ & 95.2 $\pm$ 2.2\\
\hline
$Dist.~Both~Cues$ & 69.6 $\pm$ 4.15
\end{tabular}
\vspace*{2.5cm}
\end{table*}

\newpage
\section{Conclusions}
\label{sec:conclusion}

In this paper we describe a simple experimental setup for investigating cue selection by neural networks.\\Our results show that the network favors the low-level over the high-level cue when both cues are equally present. However, when we introduce false patterns to the low-level part of the dataset, the network compensates by using also the high-level cue.\\A possible explanation why the low-level cue is preferred when both cues are equally present (experiment A, see Section~\ref{subsec:results_A}) may be the complexity of the cues. Table~\ref{tab:iterations} shows the mean number of epochs the neural networks need for converging to the final test accuracy. Learning the symbols requires by far the largest amount of epochs. Apparently, minimizing the cost function is more difficult in this case than for the other cues (all below $100$ epochs). Thus, if both cues are present in the dataset, two equally deep local minima exist, and the network will converge to the configuration corresponding the minimum that can be reached with fewer iterations. This interpretation is supported by the case of training on the subset $Dist.~Both~Cues^{train}$. If the local minimum corresponding to the pattern has a value larger than the local minimum corresponding to the symbols, the network will move towards the local minimum of the symbols, which is an absolute minimum.\\In the future, we are interested in developing strategies that allow shifting learning to specific, user defined cues. This could potentially be obtained by including information about the desired cues into the training data, e.g. by labeling relevant cues in images.
\balance

\renewcommand*{\bibfont}{\normalsize\raggedright}
\bibliographystyle{ACM-Reference-Format}
\bibliography{mybibfile}

\begin{table*}[b]
\centering
\caption{Mean test accuracies for different setups of neural networks and training datasets.}
\label{tab:all_results1}
\begin{tabular}{cccccc}
\# hidden neurons & data subset & $Both~Cues^{test}$ & $Symbol^{test}$ & $Pattern^{test}$ & $Dist.~Both~Cues^{test}$\\
\hline \\
\multicolumn{6}{c}{1 hidden layer}\\
\hline \hline \\
10 & $Both~Cues^{train}$ & \textbf{97.42 $\pm$ 0.03} & \textit{33.01 $\pm$ 0.15} & \textbf{95.81 $\pm$ 0.14} & 76.06 $\pm$ 0.31 \\
& $Symbol^{train}$ & 61.17 $\pm$ 1.50 & \textbf{99.75 $\pm$ 0.09} & \textit{37.86 $\pm$ 0.63} & 60.90 $\pm$ 1.61 \\
& $Pattern^{train}$ & 78.73 $\pm$ 0.19 & \textit{33.54 $\pm$ 0.26} & \textbf{99.25 $\pm$ 0.02} & 63.93 $\pm$ 0.24 \\
& $Dist.~Both~Cues^{train}$ & \textbf{92.63 $\pm$ 0.40} & 73.31 $\pm$ 1.76 & 81.22 $\pm$ 0.92 & 80.41 $\pm$ 0.99 \\
\\
100 & $Both~Cues^{train}$ & \textbf{97.46 $\pm$ 0.04} & \textit{33.26 $\pm$ 0.17} & \textbf{96.17 $\pm$ 0.08} & 75.61 $\pm$ 0.12 \\
& $Symbol^{train}$ & 67.93 $\pm$ 0.71 & \textbf{100 $\pm$ 0.00} & \textit{38.69 $\pm$ 0.30} & 67.95 $\pm$ 0.45 \\
& $Pattern^{train}$ & 79.18 $\pm$ 0.25 & \textit{33.48 $\pm$ 0.25} & \textbf{99.25 $\pm$ 0.02} & 64.33 $\pm$ 0.12 \\
& $Dist.~Both~Cues^{train}$ & \textbf{96.31 $\pm$ 0.16} & 85.01 $\pm$ 0.53 & 77.38 $\pm$ 0.14 & 87.09 $\pm$ 0.12 \\
\\
500 & $Both~Cues^{train}$ & \textbf{97.59 $\pm$ 0.03} & \textit{33.28 $\pm$ 0.18} & \textbf{96.54 $\pm$ 0.13} & 75.92 $\pm$ 0.29\\
& $Symbol^{train}$ & 66.33 $\pm$ 0.35 & \textbf{100 $\pm$ 0.00} & \textit{39.10 $\pm$ 0.30} & 65.91 $\pm$ 0.16 \\
& $Pattern^{train}$ & 79.10 $\pm$ 0.23 & \textit{33.35 $\pm$ 0.25} & \textbf{99.19 $\pm$ 0.02} & 63.89 $\pm$ 0.13 \\
& $Dist.~Both~Cues^{train}$ & \textbf{97.31 $\pm$ 0.10} & 86.45 $\pm$ 0.58 & 78.53 $\pm$ 0.38 & 86.53 $\pm$ 0.19 \\
\\
\multicolumn{6}{c}{2 hidden layers}\\
\hline \hline \\
10 & $Both~Cues^{train}$ & \textbf{97.14 $\pm$ 0.04} & \textit{32.95 $\pm$ 0.15} & \textbf{94.62 $\pm$ 0.80} & 75.76 $\pm$ 0.10 \\
& $Symbol^{train}$ & 60.67 $\pm$ 0.63 & \textbf{99.67 $\pm$ 0.08} & \textit{38.13 $\pm$ 0.60} & 60.49 $\pm$ 0.71 \\
& $Pattern^{train}$ & 77.39 $\pm$ 0.31 & \textit{33.81 $\pm$ 0.13} & \textbf{99.24 $\pm$ 0.01} & 62.95 $\pm$ 0.31 \\
& $Dist.~Both~Cues^{train}$ & \textbf{92.18 $\pm$ 0.48} & 58.99 $\pm$ 3.38 & 75.57 $\pm$ 1.29 & 76.91 $\pm$ 0.46 \\
\\
100 & $Both~Cues^{train}$ & \textbf{97.21 $\pm$ 0.04} & \textit{32.83 $\pm$ 0.25} & \textbf{95.10 $\pm$ 0.09} & 75.92 $\pm$ 0.20 \\
& $Symbol^{train}$ & 62.67 $\pm$ 0.63 & \textbf{99.95 $\pm$ 0.02} & \textit{38.63 $\pm$ 0.34} & 62.44 $\pm$ 0.57 \\
& $Pattern^{train}$ & 78.08 $\pm$ 0.14 & \textit{33.58 $\pm$ 0.17} & \textbf{99.19 $\pm$ 0.02} & 63.06 $\pm$ 0.06 \\
& $Dist.~Both~Cues^{train}$ & \textbf{94.43 $\pm$ 0.21} & 84.16 $\pm$ 1.13 & 67.69 $\pm$ 0.52 & 86.12 $\pm$ 0.25 \\
\\
500 & $Both~Cues^{train}$ & \textbf{97.34 $\pm$ 0.03} & \textit{33.37 $\pm$ 0.23} & \textbf{95.78 $\pm$ 0.16} & 75.77 $\pm$ 0.17 \\
& $Symbol^{train}$ & 61.34 $\pm$ 0.35 & \textbf{100 $\pm$ 0.00} & \textit{38.32 $\pm$ 0.17} & 60.99 $\pm$ 0.31 \\
& $Pattern^{train}$ & 78.27 $\pm$ 0.19 & \textit{33.55 $\pm$ 0.18} & \textbf{99.25 $\pm$ 0.02} & 63.72 $\pm$ 0.18 \\
& $Dist.~Both~Cues^{train}$ & \textbf{95.42 $\pm$ 0.16} & \textbf{93.03 $\pm$ 0.14} & 66.41 $\pm$ 0.49 & 86.32 $\pm$ 0.15 \\
\end{tabular}
\caption*{\vspace{2.5cm}}
\end{table*}

\begin{table*}[t]
\centering
\caption{Mean test accuracies for different setups of neural networks and training datasets.}
\label{tab:all_results2}
\begin{tabular}{cccccc}
\# hidden neurons & data subset & $Both~Cues^{test}$ & $Symbol^{test}$ & $Pattern^{test}$ & $Dist.~Both~Cues^{test}$\\
\hline \\
\multicolumn{6}{c}{3 hidden layers}\\
\hline \hline \\
10 & $Both~Cues^{train}$ & \textbf{96.97 $\pm$ 0.06} & \textit{33.05 $\pm$ 0.07} & \textbf{94.98 $\pm$ 0.49} & 75.42 $\pm$ 0.20 \\
& $Symbol^{train}$ & 55.98 $\pm$ 1.55 & \textbf{99.62 $\pm$ 0.13} & \textit{36.47 $\pm$ 0.80} & 56.17 $\pm$ 1.50 \\
& $Pattern^{train}$ & 78.92 $\pm$ 1.56 & \textit{33.68 $\pm$ 0.30} & \textbf{99.22 $\pm$ 0.04} & 63.32 $\pm$ 1.08 \\
& $Dist.~Both~Cues^{train}$ & \textbf{91.60 $\pm$ 0.25} & 59.85 $\pm$ 3.96 & 75.94 $\pm$ 2.75 & 76.66 $\pm$ 0.41 \\
\\
100 & $Both~Cues^{train}$ & \textbf{96.90 $\pm$ 0.05} & \textit{32.83 $\pm$ 0.22} & \textbf{95.04 $\pm$ 0.15} & 75.33 $\pm$ 0.15 \\
& $Symbol^{train}$ & 60.99 $\pm$ 0.58 & \textbf{99.86 $\pm$ 0.03} & \textit{37.66 $\pm$ 0.16} & 60.40 $\pm$ 0.58 \\
& $Pattern^{train}$ & 78.03 $\pm$ 0.28 & \textit{33.51 $\pm$ 0.25} & \textbf{99.19} $\pm$ 0.00 & 62.79 $\pm$ 0.25 \\
& $Dist.~Both~Cues^{train}$ & \textbf{92.75 $\pm$ 0.17} & 81.75 $\pm$ 0.58 & 66.54 $\pm$ 1.09 & 83.13 $\pm$ 0.47 \\
\\
500 & $Both~Cues^{train}$ & \textbf{97.00 $\pm$ 0.04} & \textit{32.89 $\pm$ 0.13} & \textbf{95.41 $\pm$ 0.23} & 75.49 $\pm$ 0.19 \\
& $Symbol^{train}$ & 57.94 $\pm$ 0.36 & \textbf{99.90 $\pm$ 0.02} & \textit{37.82 $\pm$ 0.41} & 57.56 $\pm$ 0.37 \\
& $Pattern^{train}$ & 77.76 $\pm$ 0.34 & \textit{33.22 $\pm$ 0.22} & \textbf{99.17 $\pm$ 0.01} & 63.52 $\pm$ 0.32 \\
& $Dist.~Both~Cues^{train}$ & \textbf{92.30 $\pm$ 0.28} & 87.58 $\pm$ 0.81 & 62.89 $\pm$ 0.88 & 82.09 $\pm$ 0.13 \\
\\
\multicolumn{6}{c}{10 hidden layers}\\
\hline \hline \\
10 & $Both~Cues^{train}$ & \textbf{96.42 $\pm$ 0.06} & \textit{32.81 $\pm$ 0.45} & \textbf{91.86 $\pm$ 3.26} & 75.07 $\pm$ 0.13 \\
& $Symbol^{train}$ & 44.51 $\pm$ 3.55 & \textbf{85.13 $\pm$ 12.80} & \textit{35.60 $\pm$ 0.55} & 44.31 $\pm$ 3.48 \\
& $Pattern^{train}$ & 72.23 $\pm$ 3.11 & \textit{33.18 $\pm$ 0.16} & \textbf{99.00 $\pm$ 0.03} & 58.86 $\pm$ 1.95 \\
& $Dist.~Both~Cues^{train}$ & \textbf{93.27 $\pm$ 1.08} & 39.67 $\pm$ 7.57 & 81.64 $\pm$ 3.88 & 74.86 $\pm$ 0.84 \\
\\
100 & $Both~Cues^{train}$ & \textbf{96.20 $\pm$ 0.10} & \textit{33.21 $\pm$ 0.24} & \textbf{95.90 $\pm$ 0.67} & 75.00 $\pm$ 0.26 \\
& $Symbol^{train}$ & 52.76 $\pm$ 1.70 & \textbf{98.97 $\pm$ 0.09} & \textit{37.46 $\pm$ 0.43} & 52.74 $\pm$ 1.88 \\
& $Pattern^{train}$ & 78.71 $\pm$ 1.62 & \textit{33.72 $\pm$ 0.22} & \textbf{98.66} $\pm$ 0.06 & 63.77 $\pm$ 1.01 \\
& $Dist.~Both~Cues^{train}$ & \textbf{93.92 $\pm$ 0.05} & 32.99 $\pm$ 0.21 & 93.36 $\pm$ 1.54 & 73.88 $\pm$ 0.08 \\
\\
500 & $Both~Cues^{train}$ & \textbf{96.30 $\pm$ 0.13} & \textit{33.14 $\pm$ 0.12} & \textbf{94.79 $\pm$ 0.82} & 75.12 $\pm$ 0.29 \\
& $Symbol^{train}$ & 50.31 $\pm$ 0.79 & \textbf{99.24 $\pm$ 0.05} & \textit{36.84 $\pm$ 0.42} & 49.89 $\pm$ 0.78 \\
& $Pattern^{train}$ & 78.73 $\pm$ 0.90 & \textit{33.66 $\pm$ 0.14} & \textbf{98.75 $\pm$ 0.04} & 63.67 $\pm$ 0.57 \\
& $Dist.~Both~Cues^{train}$ & \textbf{92.98 $\pm$ 0.52} & 42.45 $\pm$ 9.39 & 86.15 $\pm$ 7.88 & 75.02 $\pm$ 1.56 \\
\end{tabular}
\end{table*}

\newpage
\section*{Appendix A: All results}

\end{document}